\documentclass[letterpaper, 10 pt, conference]{ieeeconf}
\IEEEoverridecommandlockouts
\usepackage{amsmath}
\usepackage{amssymb}
\usepackage{graphicx}
\usepackage{booktabs}
\usepackage{multirow}
\usepackage[table]{xcolor}

\definecolor{oursColor}{HTML}{da5b00}
\newcommand{\ours}{decMHT}

\title{\LARGE \bf Learning Multi-Humanoid Pickup and Transport \\via Decentralized Object-Centric Control}

\author{Bikram Pandit$^{1}$, Mohitvishnu S. Gadde$^{1}$, Aayam Kumar Shrestha$^{1}$, and Alan Fern$^{1}$
\thanks{*This work is supported by NSF Award 2321851, DARPA contract HR0011-24-9-0423, and the NVIDIA Academic Grant Program.}
\thanks{$^{1}$All authors are with the Dynamics Robotics and AI Lab (DRAIL), Oregon State University, Corvallis, OR, USA. {\tt\small \{panditb, gaddem, aayam.shrestha, afern\}@oregonstate.edu}}}

\begin{document}

\maketitle
\thispagestyle{empty}
\pagestyle{empty}

\begin{figure*}[t]
    \centering
    \includegraphics[width=\textwidth]{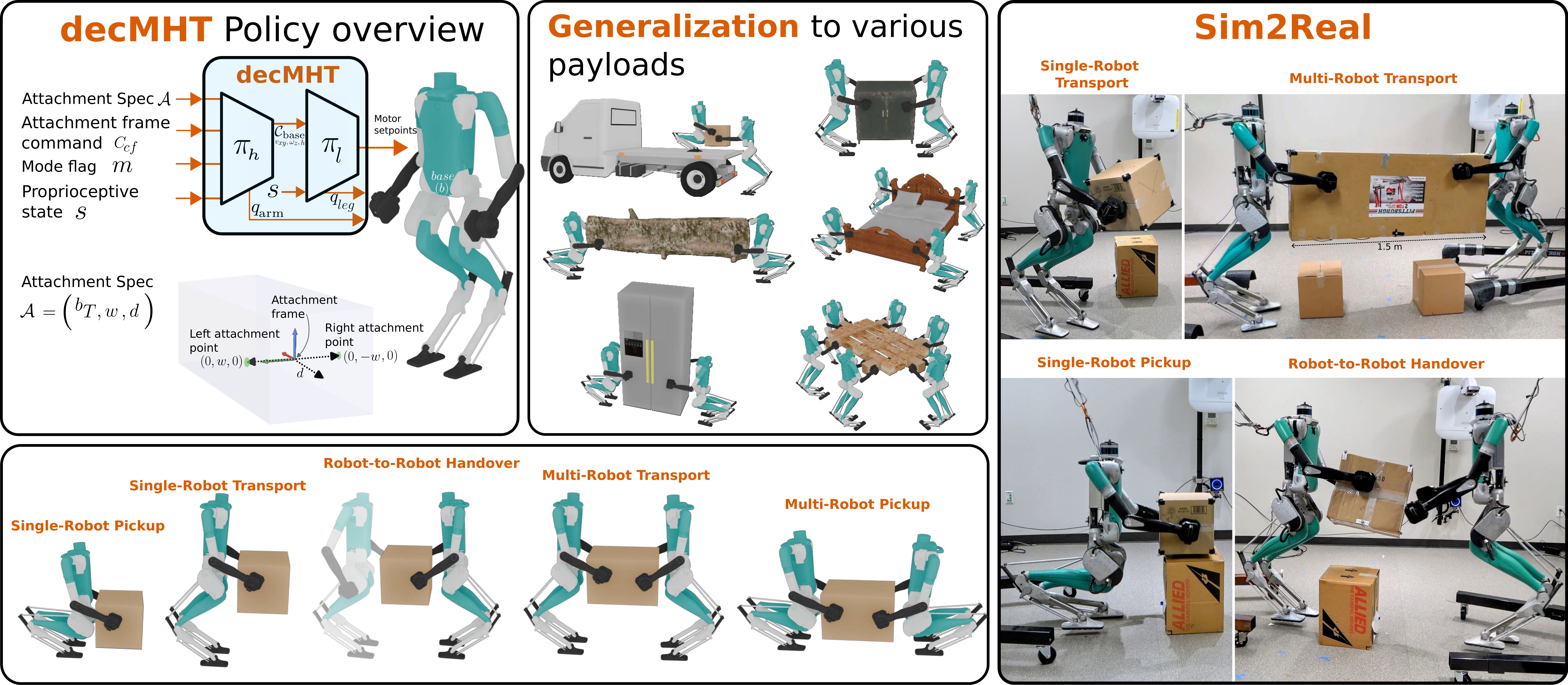}
    \caption{Our decentralized object-centric interface unifies single-robot pickup, cooperative multi-robot transport, and handover under one policy. Each robot is assigned a local attachment region on the shared object and executes the same policy independently, with no inter-robot communication, scaling from a single robot to larger teams carrying larger or more irregular payloads.}
    \label{fig:lead}
\end{figure*}

\begin{abstract}
We study cooperative multi-humanoid pickup and transport of objects with varying size, weight, and geometry, requiring robot teams of different sizes. Our approach uses decentralized object-centric control, where each humanoid is assigned a local attachment region on the shared object and learns to realize pickup and transport through gripperless bimanual pinching. This attachment-based interface provides a common control abstraction spanning single-robot pickup, cooperative multi-robot transport, and robot-to-robot handover, without per-task redesign. We find that policies trained only on single-robot pickup already transfer nontrivially to cooperative settings, suggesting that this abstraction captures much of the structure needed for coordination. At the same time, explicit multi-robot training further improves performance, showing that shared-object coupling introduces coordination dynamics that are beneficial to learn directly. We validate the approach in simulation across varying team sizes and object geometries, and demonstrate sim-to-real transfer on hardware, where the learned controllers enable real humanoids to perform cooperative manipulation tasks.
\end{abstract}

\section{INTRODUCTION}


Recent advances in humanoid loco-manipulation have enabled humanoids to pick up, carry, and place objects of varying size and weight \cite{dao2023simtoreal,barreiros2025punyo,zheng2025embracing}. However, many real-world payloads remain too large, heavy, or irregular to be manipulated by a single humanoid. Such tasks instead call for teams of robots that can coordinate their whole-body motion while interacting through a shared load. Inspired by how humans collaboratively move bulky objects, we study a learning-based framework for multi-humanoid collaborative loco-manipulation with variable-sized teams. To support this setting, we introduce \emph{decentralized multi-humanoid transport (\ours{})}, a decentralized, object-centric control abstraction. A high-level planner specifies only per-robot attachment regions and the desired object motion, while the low-level controllers autonomously realize coordinated pickup, transport, and handoff using only local observations and without direct communication. Importantly, this abstraction generalizes across varying team sizes, attachment layouts, and object geometries, while also allowing robots to join or leave the manipulation task through humanoid-to-humanoid handoffs.


Prior work on cooperative humanoid transport is still sparse. Robotics work with physically realistic robot models or real humanoids has largely focused on two-agent settings, including control-theoretic methods in simulation \cite{wen2024coordinated} and recent learning-based human-humanoid transport under leader-follower assumptions \cite{du2025cola}. A notable exception is the symmetric position/force control framework \cite{wu2016}, which demonstrates cooperative transport with up to four humanoids in dynamic simulation. However, the approach relies on analytic models of the robots and manipulated object and does not demonstrate a learned controller that generalizes across team sizes, attachment layouts, or object geometries. To our knowledge, our work is the first to demonstrate learned, decentralized multi-humanoid loco-manipulation across varying team sizes in robot-realistic simulation, together with deployment on real humanoid hardware through a common object-centric control abstraction.

In summary our main contributions are fourfold:

\begin{itemize}
\item We introduce \ours{}, a high-level, object-centric abstraction for multi-humanoid loco-manipulation that supports variable team sizes, attachment layouts, and object geometries.

\item We develop a hierarchical decentralized control architecture, together with a training curriculum and reward design, that enables humanoids to coordinate through the shared object using only local observations and without direct communication.

\item We demonstrate in simulation that a single trained controller supports teams of up to ten humanoids across diverse object geometries and attachment configurations, including humanoid-to-humanoid handoffs across multiple elevation differences.

\item We demonstrate sim-to-real transfer of the approach on two Digit V3 humanoids performing collaborative object manipulation.
\end{itemize}

\section{RELATED WORK}
\label{sec:related}


\textbf{Single-Robot Humanoid Loco-Manipulation.}
Recent work has shown that single humanoids can perform contact-rich pickup and transport of large or heavy payloads through reinforcement learning. Some examples include sim-to-real box loco-manipulation on Digit across varying payload weights, poses, and orientations \cite{dao2023simtoreal}; example-guided learning for lifting and stowing bulky objects with a compliant humanoid torso \cite{barreiros2025punyo}; motion-prior-based whole-body control for embracing and transporting bulky objects \cite{zheng2025embracing}; and imitation of
dynamically-feasible trajectories from trajectory optimization for contact-rich
loco-manipulation \cite{Liu2024Opt2SkillID}. Others push single-robot capacity further through residual learning on a motion-tracking prior \cite{Zhao2025ResMimicFG}, factorized load adaptation \cite{Kang2026SplitAdapterLH}, and test-time planning \cite{Zhang2026SumoDA}. Together, these works demonstrate gripperless whole-body manipulation, but remain limited by the strength and reach of a single robot.


\textbf{Model-Based Multi-Robot Transport.} Control-theoretic and model-based approaches have coordinated multiple legged robots carrying a shared payload using MPC, impedance control, and distributed optimization \cite{turrisi2024pacc,zhou2026aclm,wen2024coordinated,Hawley2019ControlFF, Wang2025SharedOM, Vincenti2023CentralizedMP}. These methods generally assume known coupled robot-payload dynamics and pre-specified mechanical contacts, and are demonstrated with small, fixed team sizes. In contrast, our approach is model-free, requires no shared dynamics model or pre-specified contact pose, and generalizes across unseen team sizes and attachment layouts.

\textbf{Decentralized Learning-Based Cooperative Transport.}
Closest to our work are decentralized learning-based methods for cooperative transport. decMBC \cite{pandit2025decmbc} learns a shared controller for arbitrarily sized teams of bipeds rigidly attached to a carrier, while decPLM \cite{pandit2026decplm} enables gripperless pinch-lift-move behavior for quadruped-arm teams under fixed, externally specified contact frames. DeReCo \cite{shibata2026dereco} targets object-geometry generalization for pairs of wheeled robots. PAINT \cite{cao2026paint} and COLA \cite{du2025cola} instead focus on compliant interaction with a partner, using proprioceptive feedback to infer or respond to partner intent without explicit communication. Force-based coordination has also been demonstrated for two wheeled robots manipulating long objects, but requires instrumented grasping hardware \cite{BernardTiong2024CooperativeGA}. These works establish that communication-free cooperative transport is possible, but they remain tied to non-humanoid embodiments, compliant-partner settings, specialized sensing, or a single manipulation mode. In contrast, decMHT provides a common attachment-spec interface for multi-humanoid pickup, cooperative transport, and handover, while generalizing across team sizes, attachment layouts, and object geometries with hardware validation.


\textbf{Collaborative Animated Humanoid Characters.}
CooHOI \cite{gao2024coohoi} and TeamHOI \cite{lionar2026teamhoi} study cooperative object transport with physics-based animated humanoid characters. Like decMHT, they seek flexible policies that generalize across cooperative configurations, with TeamHOI supporting variable team sizes through a shared transformer policy. However, these methods operate on animated characters rather than robot-realistic humanoid models, and do not address whole-body robot control or hardware transfer. In contrast, our work targets physically realistic humanoid robot models and demonstrates the same type of flexibility across team sizes and object geometries, together with deployment on real hardware.

\section{Decentralized Transport Framework}
\label{sec:method}


We consider decentralized cooperative pickup and transport in which a team of N humanoid robots must approach, lift, and transport a shared payload using gripperless bimanual contact, with optional handoffs between robots. Our goal is to provide a simple object-centric interface through which a high-level planner specifies only the information needed for each robot to contribute to the task. The interface has two components: an \emph{attachment spec}, provided once at the start of a manipulation episode to define each robot’s local attachment region, and an \emph{object-centric command}, provided at each timestep to specify the desired payload motion and transformed into a corresponding local command for each robot through a fixed rigid transformation. Given this information, each robot independently executes a local controller using only its interface inputs and local observations. Unlike prior work \cite{pandit2025decmbc}, which assumes rigid mechanical attachment, and \cite{pandit2026decplm}, which assumes a fixed contact pose specified by an external planner, our formulation avoids precise robot-level pose commands while remaining composable across robots and object geometries.

\textbf{Attachment-Spec Representation.}
The \emph{attachment spec} describes approximately where a robot should attach to the payload without prescribing a single fixed contact pose. It provides the local geometric information needed for the controller to locomote as needed, avoid collisions with the payload, and establish flush bimanual contact.

Formally, for robot $r$, we define the attachment spec as
$\mathcal{A}^{(r)} = \left({}^{b}T^{(r)}, w^{(r)}, d^{(r)}\right)$,
where ${}^{b}T^{(r)} \in SE(3)$ specifies the pose of the attachment frame relative to the robot's torso frame $b$. The scalar $w^{(r)}$ specifies symmetric left and right attachment points at $(0,w^{(r)},0)$ and $(0,-w^{(r)},0)$, respectively, expressed in the attachment frame. The scalar $d^{(r)}$ specifies the clearance from the attachment-frame origin toward the robot-facing boundary of the payload, providing the geometric information needed to position the torso during approach. Since the payload is rigid, $w^{(r)}$ and $d^{(r)}$ are fixed for a given attachment region and need only be measured once. In contrast, the attachment-frame pose ${}^{b}T^{(r)}$ must be tracked online as the robot and payload move relative to one another. In simulation, this pose is obtained directly from the object state, while on hardware it is estimated using an external sensor.


Critically, the attachment spec does not attempt to describe the full geometry of the payload’s attachment region. It specifies only the two intended attachment points and the frontal clearance needed for approach, while leaving the payload extent above, below, and beyond the attachment region unspecified. This allows the same representation to describe both compact objects, such as boxes, and elongated or irregular objects, such as bars and panels, whose global geometry may vary substantially or extend to attachment regions assigned to other robots. Any object that admits approximately flat opposing contact surfaces can be represented by placing an attachment frame between the intended hand-contact points and specifying the local frontal clearance. Thus, the attachment spec captures the local geometry needed by each robot without requiring a complete geometric model of the attachment region.

\textbf{Object-Centric Command.}
A task-level object-centric command
$\mathcal{C}_{\mathrm{pl}} = (v_x, v_y, \omega_z, h)$
specifies the desired planar velocity, yaw rate, and height of a user-chosen reference frame on the object, such as its centroid. For each robot $r$, this command is transformed into a local command
$\mathcal{C}^{(r)}_{\mathrm{cf}} =
\left(v_x^{(r)}, v_y^{(r)}, \omega_z^{(r)}, h^{(r)}\right)$
defined at the robot's attachment-spec frame. Letting
$p^{(r)}_{\mathrm{offset}}$
denote the fixed offset from the payload reference frame to robot $r$'s local command frame, the transformed command is
\begin{small}
\begin{align*}
    v^{(r)}_{\mathrm{cf}} &= v_{\mathrm{pl}} + \omega_{\mathrm{pl}} \times p^{(r)}_{\mathrm{offset}}, \\
    \omega^{(r)}_{\mathrm{cf}} &= \omega_{\mathrm{pl}}, \qquad h^{(r)}_{\mathrm{cf}} = h_{\mathrm{pl}} + p^{(r)}_{\mathrm{offset},z},
\end{align*}
\end{small}
where $\omega_{\mathrm{pl}} = \omega_z\,\hat{z}$ is the commanded yaw rate expressed as a vector about the object's vertical axis. Because the payload is rigid,
$p^{(r)}_{\mathrm{offset}}$ is constant for a given attachment assignment and does not change as the payload moves. Thus, a single object-level command can be converted cheaply into the corresponding local command for each robot using a fixed rigid transformation.



Each robot additionally receives a binary mode flag $m^{(r)} \in \{\texttt{idle}, \texttt{track}\}$.
In \texttt{idle} mode, the robot maintains a fixed nominal pose, while in \texttt{track} mode it approaches, attaches to, and tracks $\mathcal{C}^{(r)}_{\mathrm{cf}}$ using its assigned attachment spec. The flag is set externally for each robot and supports pickup and transport by placing the relevant robots in \texttt{track}, placement by switching them to \texttt{idle}, and handover by first activating the receiving robot and then deactivating the releasing robot once contact is established. The transition timing is specified externally, and each local robot controller is aware only of its own mode, not the modes or states of other robots.

\section{Decentralized Control Architecture}
\label{sec:system}



Each robot executes the same control policy using its robot-specific attachment spec and object-centric command. We adopt a two-level hierarchical architecture in which a high-level task policy produces upper-body joint targets and locomotion commands, while a low-level locomotion policy realizes the commanded base motion.


\textbf{Low-Level Locomotion Policy.}
The low-level policy $\pi_l$ provides a velocity-, yaw-rate-, and height-control abstraction over the robot's legs:
\begin{equation*}
    q^{(r)}_{\mathrm{leg}} = \pi_l\left(s^{(r)}, v_x^{(r)}, v_y^{(r)}, \omega_z^{(r)}, h^{(r)}\right),
\end{equation*}
where $s^{(r)}$ denotes the proprioceptive state, consisting of joint positions and velocities, IMU orientation, and angular velocity. The inputs $v_x^{(r)}$, $v_y^{(r)}$, $\omega_z^{(r)}$, and $h^{(r)}$ specify the commanded planar velocity, yaw rate, and torso height, respectively. The output $q^{(r)}_{\mathrm{leg}}$ consists of target leg-joint positions tracked by PD controllers. The low-level locomotion policy $\pi_l$ is a neural network that is pretrained separately using the multi-modal whole-body control approach~\cite{dugar2025learning} and remains frozen during high-level policy training.

\textbf{High-Level Manipulation Policy.} The high-level policy $\pi_h$ coordinates arm motion and generates locomotion targets to realize approach, contact, lifting, and command tracking:
\begin{small}
\begin{equation*}
    \left(
    q^{(r)}_{\mathrm{arm}},
    v_x^{(r)},
    v_y^{(r)},
    \omega_z^{(r)},
    h^{(r)}
    \right)
    =
    \pi_h\left(
    s^{(r)},
    \mathcal{A}^{(r)},
    \mathcal{C}^{(r)}_{\mathrm{cf}},
    m^{(r)}
    \right),
\end{equation*}
\end{small}
where $q^{(r)}_{\mathrm{arm}}$ denotes target upper-body joint positions, and $\left(v_x^{(r)}, v_y^{(r)}, \omega_z^{(r)}, h^{(r)}\right)$ is passed to the low-level locomotion policy. The output yaw rate $\omega_z^{(r)}$ commands the robot's base and need not equal the object-command yaw rate $\omega_{\mathrm{cf}}^{(r)}$; for example, the robot may sidestep to realize the commanded object rotation while maintaining a comparatively fixed heading. Through this common output interface, $\pi_h$ can walk toward a distant payload, squat to reach a low attachment region, and track $\mathcal{C}^{(r)}_{\mathrm{cf}}$ after contact is established. The policy $\pi_h$ is implemented as a two-layer LSTM with a 128-dimensional hidden state per layer, similar to the recurrent policies used for locomotion control in~\cite{siekmannsim2real}, to support history-dependent contact behavior.



\textbf{Decentralized Execution.} At runtime, each robot independently executes $\pi_h$ and $\pi_l$ at 50\,Hz using only its local observations, attachment spec, object-centric command, and mode flag, with no communication between robots. Handover is coordinated externally through the mode flags.

\section{Training Approach}

We train the high-level policy $\pi_h$ via reinforcement learning using Independent Proximal Policy Optimization (IPPO) \cite{schulman2017ppo, yu2022ippo} run on simulated episodes initialized from randomized scene configurations and a staged curriculum that progressively increases the difficulty of contact formation, payload support, command tracking, and locomotion. Training first focuses on developing robust single-robot pickup and transport capabilities, which provide the basic skills needed by every robot in the decentralized system. The resulting policy is then fine-tuned using two-robot episodes to expose it to the coupled dynamics of cooperative manipulation. Although training never uses more than two robots, we evaluate both the single-robot-trained and two-robot-fine-tuned policies zero-shot with teams of up to ten robots.

\subsection{Initial Scene Generation}

All training episodes use cuboid payloads with randomized dimensions and masses. These boxes provide a simple means of generating diverse attachment specs and are intended to represent local attachment regions that may belong to substantially larger or non-box-shaped objects. For single-robot training, box dimensions range from $0.1 \times 0.1 \times 0.1$\,m to $0.7 \times 0.5 \times 0.7$\,m. For two-robot training, the minimum dimensions are increased to $0.3 \times 0.1 \times 0.1$\,m to provide sufficient space for both robots. Payload mass is sampled from $0.5$ to $1.5$\,kg.

The attachment frame is sampled over half of the graspable face area facing the robot, with $0.1$\,m padding from the face boundaries. The object-centric command frame is sampled with an offset of up to $0.5$\,m from the attachment frame in both $x$ and $y$, together with full yaw coverage about the vertical axis. Each robot is initialized relative to its assigned attachment region with longitudinal offset $x \in [-1.0,-0.2]$\,m, lateral offset $y \in [-0.5,0.5]$\,m, and yaw offset in $[-10^\circ,10^\circ]$.

Objects are initialized at randomized heights using an external support force, allowing the policy to practice contact formation at elevations that would otherwise be infeasible before grasping. The support is removed after contact is detected over two consecutive $0.05$\,s intervals, after which the robot must support the payload under normal simulated dynamics. The support may be re-enabled when the commanded mode requires the robot to release the payload, permitting multiple pickup, release, and handoff transitions within a single episode.

\subsection{Training Curriculum}

The curriculum first develops the single-robot capabilities needed for approach, bimanual attachment, lifting, and object-command tracking. Only after these skills have converged do we introduce multi-robot training, so that cooperative learning begins from a controller that already knows how to manipulate its assigned local attachment region.

\textbf{Stage 1 -- Assisted Contact Formation.}
The robot is initialized near the payload with randomized relative position, orientation, attachment frame, and payload height. The payload remains externally supported and resistant to disturbance, allowing the robot to learn approach and flush bimanual contact without simultaneously needing to support its weight. Object-command tracking is disabled.

\textbf{Stage 2 -- Unassisted Holding.}
External support is removed after contact is established, and the robot must support and stabilize the payload under normal physics. Object-command tracking remains disabled.

\textbf{Stage 3 -- Height Tracking.}
Payload-height commands are enabled while planar velocity and yaw-rate commands remain zero. This stage teaches the robot to lift and vertically reposition the payload while maintaining contact.

\textbf{Stage 4 -- Full Object-Command Tracking.}
Planar velocity and yaw-rate commands are added, requiring the robot to track the full object-centric command while maintaining a stable hold.

\textbf{Stage 5 -- Distant Spawning.}
The robot is initialized farther from the payload and must locomote toward its assigned attachment region before establishing contact and tracking the full command. This stage combines approach locomotion, contact formation, lifting, and transport in a single episode.

\textbf{Stage 6 -- Two-Robot Fine-Tuning.}
The single-robot policy is fine-tuned in two-robot cooperative episodes. Each robot independently receives its own attachment spec, local object-centric command, mode flag, and local observations. This stage exposes the policy to forces generated by another robot through the shared payload while preserving the same decentralized controller and interface used during single-robot training.

During single-robot training, the mode is resampled every $3$--$8$\,s, with equal probability of \texttt{track} and \texttt{idle}. During two-robot training, the joint mode configuration is resampled over the same interval, with probability $0.5$ that both robots are in \texttt{track} mode and probability $0.25$ for each configuration in which one robot tracks and the other is idle. These transitions expose the policy to cooperative pickup and transport, solo manipulation, release, and both directions of robot-to-robot handoff within a common training process. Episodes last $10$\,s through Stage~4 and $15$\,s for Stages~5 and~6.

\subsection{Reward Design}




The reward for robot $r$ is gated by its sampled mode flag $m^{(r)}$, allowing a single policy to learn distinct idle and track behaviors.

\textbf{Idle Reward.}
When $m^{(r)} = \texttt{idle}$, the reward encourages zero
base velocity, a fixed nominal torso height, and nominal
rest-pose arm joint positions.

\textbf{Track Reward.}
When $m^{(r)} = \texttt{track}$, the reward is composed of
four terms:
\begin{enumerate}
    \item \emph{Approach}: reduces the position and orientation
    error between the robot torso and its assigned attachment
    frame until both fall within a threshold range.

    \item \emph{Contact}: once the approach criterion is met,
    reduces the distance between each hand end effector and
    its corresponding attachment point, while aligning the
    hand surface normals with the opposing contact normals to
    encourage flush contact.

    \item \emph{Holding}: encourages the object to remain in
    a target region relative to the robot torso. For low initial
    payload heights, this term induces the robot to squat,
    establish contact, and lift.

    \item \emph{Command Tracking}: once the object is held,
    reduces the error between its commanded and achieved
    planar velocity, yaw rate, and height.
\end{enumerate}

\section{Training Details and Domain Randomization.}
Full implementation details are provided in the extended paper and code release. All training is conducted in IsaacLab~\cite{mittal2023orbit} using the Agility Robotics Digit V3 humanoid with customized end effectors consisting of rigid, flat, rubberized contact pads with mechanical tolerance for small in-plane rotation errors, allowing flush contact with a target face despite imperfect alignment (Figure~\ref{fig:lead}). We train using IPPO with 4096 parallel environments for single-robot training, requiring approximately 140 GPU-hours on a single NVIDIA RTX 4090 GPU with 24\,GB of memory. Two-robot fine-tuning uses 2048 parallel environments and requires approximately 60 additional GPU-hours. The actor, critic, and optimizer state from the converged single-robot policy are used to initialize Stage~6. The policy is never trained with $N \geq 3$ robots. We refer to the resulting two-robot-fine-tuned policy as \ours{} and retain the single-robot-trained policy as a zero-shot ablation. All evaluations with $N \geq 3$ use these policies without additional fine-tuning.

To improve robustness and support sim-to-real transfer, we randomize joint-position sensor noise, IMU noise, foot-ground friction, joint and motor stiffness, and motor-response delay throughout training. Attachment-spec observations are updated at a reduced rate to model hardware sensing latency, and the payload is subjected during transport to random external velocity perturbations sampled from $[-0.5,0.5]$\,m/s.

\section{Experiments and Results}
\label{sec:experiments}


We evaluate \ours{} primarily in simulation, with hardware validation, to answer six questions: (1) How does performance scale with team size? (2) Does the attachment-based abstraction generalize to unseen object geometries and masses? (3) How does transport efficiency scale with team size? (4) What cooperative behavior emerges from single-robot training, and what improves with multi-robot fine-tuning? (5) How does \ours{} compare with centralized and object-specialized baselines? (6) Does the controller transfer to real humanoid hardware?

\subsection{Experimental Setup}

We evaluate teams of $N \in \{1,\dots,10\}$ humanoid robots performing cooperative pickup, transport, and handover in simulation. Each configuration is evaluated over 1000 episodes with randomized initial robot poses, attachment assignments, and commanded object trajectories. Pickup-and-transport episodes last 8\,s, while handover episodes last 12\,s. Commands include planar velocities up to $0.6$\,m/s, yaw rates up to $0.3$\,rad/s, and object heights in $[0.4,0.95]$\,m. 

We report:
\begin{itemize}
    \item \textbf{Pickup success rate (\%):} fraction of episodes in which the object is successfully lifted.
    \item \textbf{Linear and angular tracking error:} RMS error between commanded and achieved object velocities.
    \item \textbf{Height tracking error:} RMS error relative to the commanded object height.
    \item \textbf{Object tilt:} RMS and MAX object tilt during transport.
    \item \textbf{Drop rate (\%):} fraction of episodes in which the object is dropped or attachment contact is lost.
\end{itemize}

All simulation evaluations use the same dynamics randomization and observation noise applied during training.

\subsection{Scalability Across Team Sizes}
\label{sec:scalability}

\begin{figure*}[t]
    \centering
    \includegraphics[angle=90, width=\textwidth]{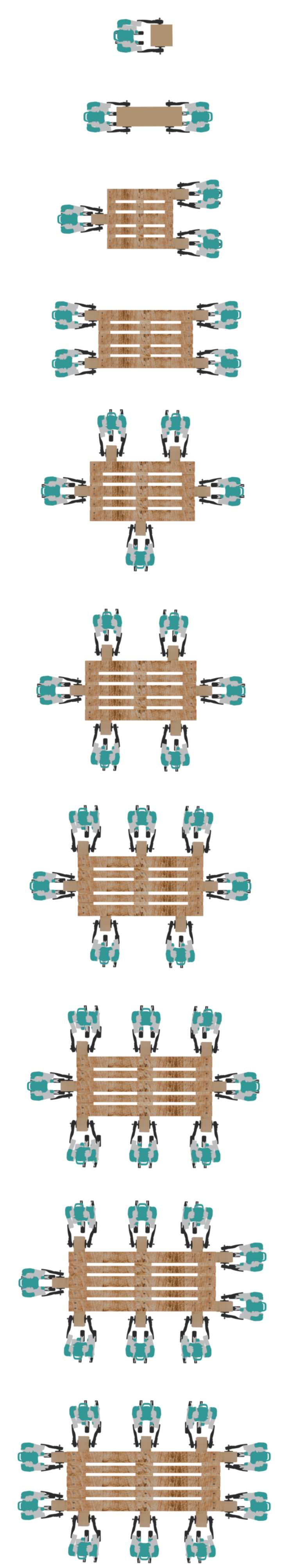}
    \caption{Top-view robot arrangements and attachment regions for teams of 1--10 humanoids.}
    \label{fig:team_configs}
\end{figure*}

We evaluate teams of $N \in \{1,\dots,10\}$ robots evenly distributed around a pallet payload (Fig.~\ref{fig:team_configs}). Object dimensions are adjusted to maintain at least $0.8$\,m between adjacent attachment points, and payload mass scales linearly from 1\,kg at $N=1$ to 10\,kg at $N=10$ to keep per-robot loading approximately constant. Each trial requires the robots to approach and pick up the payload before tracking the commanded object motion. 
Table~\ref{tab:scalability} reports performance across hold-still, forward, lateral, and turning commands.

\begin{table}[t]
    \centering
    \footnotesize
    \setlength{\tabcolsep}{3pt}
    \caption{Pickup success, tracking error, drop rate, and object tilt across team sizes for the primary box object.}
    \label{tab:scalability}
    \resizebox{\columnwidth}{!}{%
    \begin{tabular}{c|cccc c cc}
        \toprule
        $N$ & \shortstack{Pickup\\(\%)} & \shortstack{Lin.\ err.\\(m/s)} & \shortstack{Ang.\ err.\\(rad/s)} & \shortstack{Height err.\\(m)} & \shortstack{Drop\\(\%)} & \shortstack{RMS Tilt\\(deg)} & \shortstack{Max Tilt\\(deg)} \\
        \midrule
        1  & 99.2 & 0.253 & 0.193 & 0.090 & 0.6 & 8.04 & 13.26 \\
        2  & 99.5 & 0.159 & 0.097 & 0.145 & 0.0 & 2.94 & 5.68 \\
        3  & 99.5 & 0.151 & 0.082 & 0.103 & 5.5 & 3.90 & 8.10 \\
        4  & 99.9 & 0.162 & 0.106 & 0.098 & 1.4 & 3.20 & 6.70 \\
        5  & 99.6 & 0.152 & 0.122 & 0.087 & 1.5 & 2.81 & 5.13 \\
        6  & 98.8 & 0.142 & 0.143 & 0.081 & 1.5 & 2.50 & 4.55 \\
        7  & 98.5 & 0.138 & 0.151 & 0.075 & 1.5 & 2.01 & 3.47 \\
        8  & 97.3 & 0.130 & 0.147 & 0.090 & 0.9 & 1.95 & 3.33 \\
        9  & 98.3 & 0.112 & 0.097 & 0.056 & 0.5 & 1.33 & 2.08 \\
        10 & 98.3 & 0.123 & 0.084 & 0.038 & 0.5 & 1.21 & 1.45 \\
        \bottomrule
    \end{tabular}%
    }
\end{table}

Pickup success remains above 97\% for every team size, demonstrating reliable zero-shot scaling to ten robots. Transport quality also generally improves with team size: linear and height tracking errors decrease, drop rates remain low, and payload tilt falls substantially. In particular, RMS tilt decreases from $8.04^\circ$ for one robot to $1.21^\circ$ for ten, suggesting that additional attachment points improve load sharing and constrain unwanted payload rotation.

\subsection{Object Mass vs. Team Size}
\label{sec:mass_scalability}

We next evaluate generalization to payload masses far beyond those seen during training. Using the same team configurations as above, we sweep $N \in \{1,\dots,10\}$ across fixed payload masses of 10, 20, 30, 40, and 50\,kg.

\begin{figure*}[t]
    \centering
    \includegraphics[width=\textwidth]{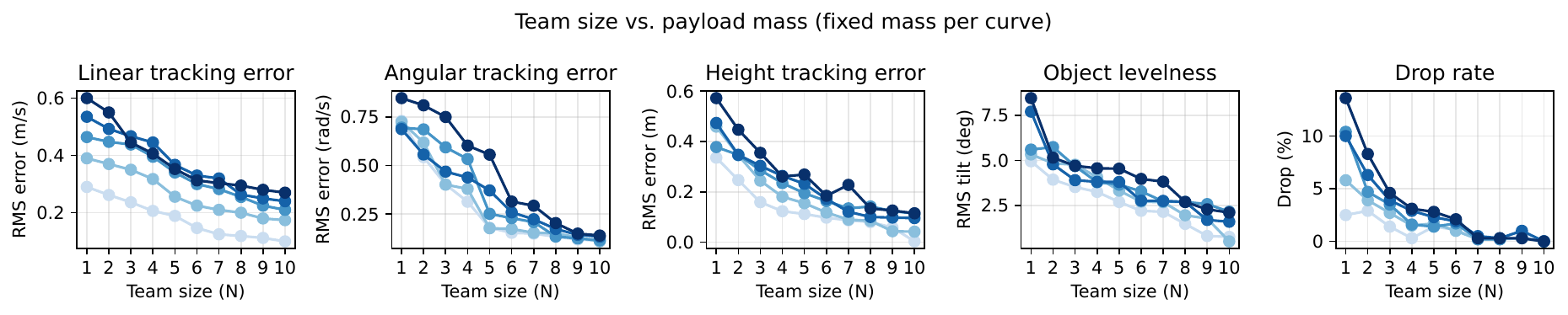}
    \caption{Tracking error, object tilt, and drop rate versus team size for fixed payload masses from 10 to 50\,kg. Darker curves indicate heavier payloads.}
    \label{fig:mass_team_size}
\end{figure*}

Figure~\ref{fig:mass_team_size} shows that increasing team size expands the range of payload masses that can be lifted and transported. A single robot handles only the lightest payload, whereas heavier payloads become feasible as additional robots share the load, with tracking and stability generally improving as the team grows. Although the policy is trained with at most two robots and much lighter payloads, it generalizes zero-shot to teams of up to ten robots carrying payloads as heavy as 50\,kg. This result suggests that the decentralized attachment-based interface enables load capacity to scale with team size without retraining for each team configuration or mass.

\subsection{Generalization to Object Geometry and Baselines}
\label{sec:generalization}

We evaluate generalization across five object geometries motivated by real-world manipulation tasks: a metal cabinet, fridge, wooden log, bed, and concrete slab. These objects vary substantially in size, shape, mass, and required team configuration, and none is used during decMHT training.

We compare \ours{} against one ablation and two baselines that are retrained separately for each evaluation object. The \emph{\ours{}(1)} ablation denotes the policy after single-robot training but before two-robot fine-tuning, and measures how much multi-robot capability emerges directly from the single-robot attachment-based abstraction. For each object, the \emph{centralized} baseline trains a single policy conditioned on the joint state of all robots, while the \emph{object-specialized} baseline trains the same decentralized architecture as \ours{} specifically on that object's geometry. The centralized baseline is omitted for the four-robot bed task because of the substantially larger joint observation and action space, which made training impractical. Table~\ref{tab:geometry} reports pickup success, tracking error, drop rate, and object tilt for all four variants across every object.
\begin{table}[t]
    \caption{Evaluation metrics across unseen object geometries, comparing \ours, the zero-shot single-robot-trained policy $\ours$(1), and centralized and specialized baselines trained per object.}
    \centering
    \footnotesize
    \setlength{\tabcolsep}{3pt}
    \renewcommand{\arraystretch}{1.4}
    \label{tab:geometry}
    \resizebox{\columnwidth}{!}{%
    \begin{tabular}{c|l|c|ccccc c c}
        \toprule
        \scriptsize Scene & \scriptsize Object & \scriptsize Training & \scriptsize\shortstack{Pickup\\(\%)} & \scriptsize\shortstack{Lin.\\(m/s)} & \scriptsize\shortstack{Ang.\\(rad/s)} & \scriptsize\shortstack{Height\\(m)} & \scriptsize\shortstack{Drop\\(\%)} & \scriptsize\shortstack{RMS T.\\(deg)} & \scriptsize\shortstack{Max T.\\(deg)} \\
        \midrule
        \multirow{4}{*}{\raisebox{-0.5\height}{\includegraphics[height=12mm]{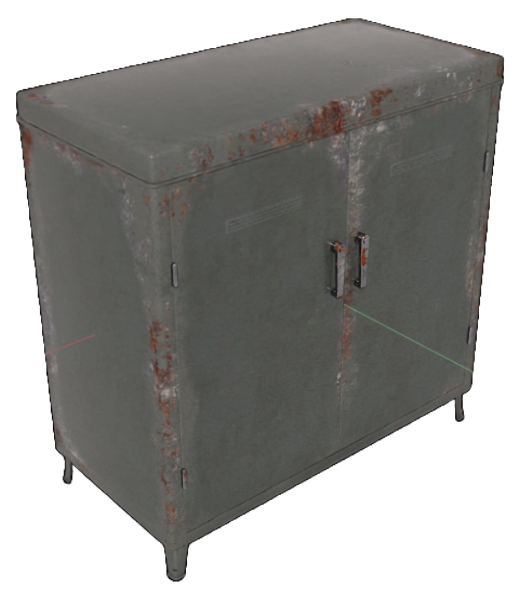}}} & \multirow{4}{*}{\scriptsize\shortstack[l]{Cabinet\\20 kg\\$N{=}2$}} & \ours{}        & 97.2  & 0.344 & 0.376 & 0.110 & 2.0 & 8.48  & 21.04 \\ 
         & & \ours(1)       & 96.6  & 0.895 & 1.115 & 0.104 & 6.9 & 23.62 & 64.22 \\
         & & Centralized  & 97.8  & 0.383 & 0.464 & 0.115 & 2.2 & 8.66  & 20.73 \\
         & & Specialized  & 96.5  & 0.394 & 0.493 & 0.110 & 4.4 & 8.02  & 18.86 \\
        \midrule
        \multirow{4}{*}{\raisebox{-0.5\height}{\includegraphics[height=14mm]{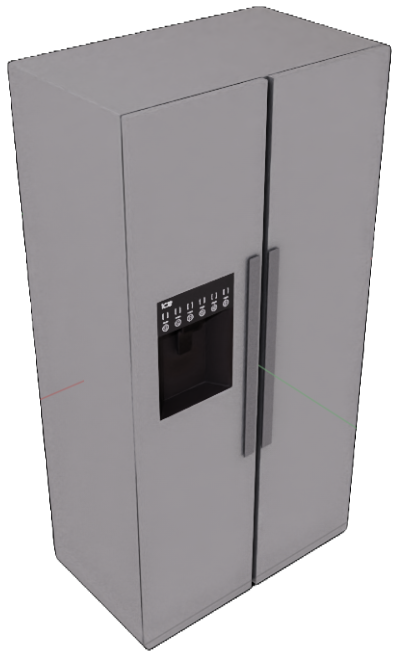}}} & \multirow{4}{*}{\scriptsize\shortstack[l]{Fridge\\20 kg\\$N{=}2$}} & \ours{}         & 95.4  & 0.266 & 0.157 & 0.400 & 4.6 & 1.77  & 3.88 \\  
         & & \ours(1)       & 96.5  & 0.424 & 0.295 & 0.401 & 5.1 & 3.84  & 11.93 \\
         & & Centralized  & 96.6  & 0.188 & 0.094 & 0.373 & 4.5 & 0.90  & 2.10 \\
         & & Specialized  & 95.9  & 0.196 & 0.110 & 0.404 & 4.9 & 1.40  & 3.26 \\
        \midrule
        \multirow{4}{*}{\raisebox{-0.5\height}{\includegraphics[height=8mm]{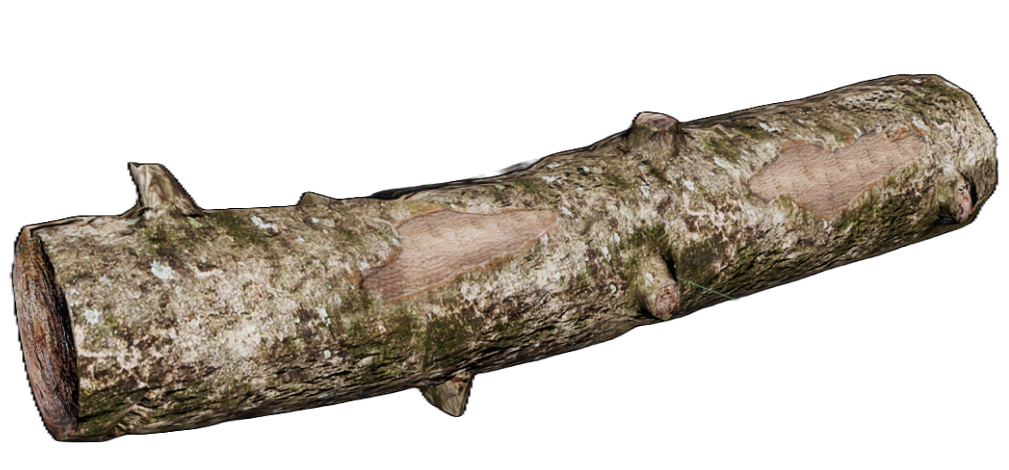}}} & \multirow{4}{*}{\scriptsize\shortstack[l]{Log\\20 kg\\$N{=}2$}} & \ours{}  & 100.0 & 0.184 & 0.095 & 0.260 & 0.5 & 11.39 & 14.92 \\
         & & \ours(1)       & 100.0 & 0.205 & 0.127 & 0.155 & 0.3 & 11.79 & 16.70 \\
         & & Centralized  & 99.9  & 0.196 & 0.100 & 0.266 & 0.0 & 10.79 & 14.39 \\
         & & Specialized  & 100.0 & 0.184 & 0.096 & 0.258 & 0.0 & 11.47 & 15.27 \\
        \midrule
        \multirow{3}{*}{\raisebox{-0.5\height}{\includegraphics[height=10mm]{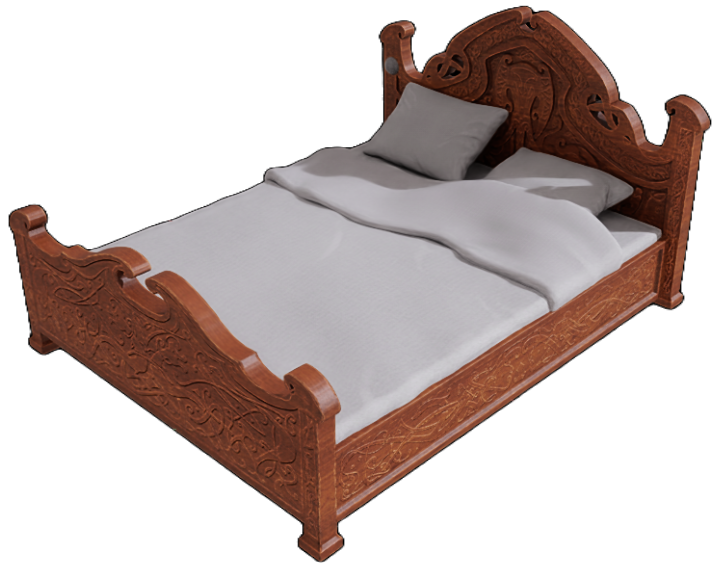}}} & \multirow{3}{*}{\scriptsize\shortstack[l]{Bed\\40 kg\\$N{=}4$}} & \ours{}         & 85.2  & 0.390 & 0.197 & 0.402 & 0.4 & 4.11  & 7.03 \\ 
         & & \ours(1)      & 44.1  & 0.603 & 0.384 & 0.327 & 0.5 & 5.56  & 8.07 \\
         & & Centralized  & --  & -- & -- & -- & -- & --  & -- \\
         & & Specialized  & 80.8  & 0.428 & 0.215 & 0.401 & 0.4 & 4.48  & 7.49 \\
        \midrule
        \multirow{4}{*}{\raisebox{-0.5\height}{\includegraphics[height=10mm]{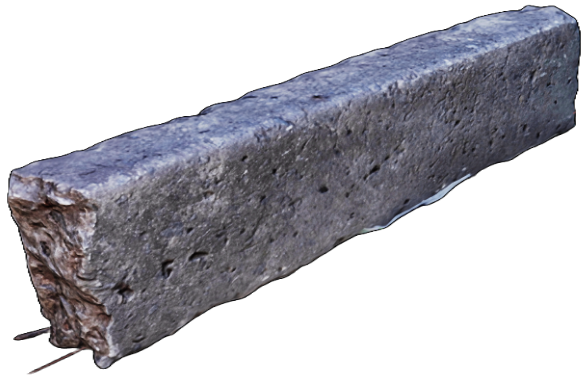}}} & \multirow{4}{*}{\scriptsize\shortstack[l]{Concrete\\30 kg\\$N{=}2$}} & \ours{}         & 100.0 & 0.222 & 0.142 & 0.315 & 0.0 & 5.39  & 13.25 \\ 
         & & \ours(1)       & 98.4  & 0.830 & 0.683 & 0.353 & 4.9 & 14.07 & 36.62 \\
         & & Centralized  & 99.9  & 0.221 & 0.164 & 0.308 & 0.1 & 5.57  & 13.61 \\
         & & Specialized  & 99.9  & 0.244 & 0.165 & 0.321 & 0.2 & 6.30  & 15.23 \\
        \bottomrule
    \end{tabular}%
    }
    \renewcommand{\arraystretch}{1}
    \vspace{-1em}
\end{table}

\textbf{\ours{} Performance.} \ours{} achieves high pickup success across all objects despite none appearing during training, demonstrating transfer to unseen geometries and masses without per-object retraining. The bed is the most challenging case, with an 85\% pickup rate, likely due to its high mass and thin protruding grasp regions for the four-robot team.

Object-specific trends also show why tilt should be considered alongside tracking error. The log achieves the lowest velocity-tracking error despite its curved contact surface, indicating robustness to imperfectly flush contact, but exhibits relatively high tilt because the rounded geometry provides little resistance to rolling. The cabinet similarly shows substantial peak tilt, consistent with its tall, top-heavy geometry. The fridge and bed also exhibit larger height-tracking errors than the lower-profile objects, suggesting that height control becomes more difficult when the commanded reference frame lies farther from the robots' nominal height.

\textbf{Comparison with \ours{}(1).} The single-robot-trained ablation achieves pickup rates comparable to \ours{} on most objects, showing that the attachment-based representation learned from single-robot training already transfers well to multi-robot pickup. The main exception is the bed, where pickup success drops from 85.2\% for \ours{} to 44.1\% for \ours{}(1). More broadly, the ablation has consistently higher tracking error, tilt, and drop rate, indicating that single-robot training captures attachment and lifting but not the coupled dynamics required for high-quality transport.

\textbf{Comparison with Object-Specific Training.} Despite receiving no training on the evaluation objects, \ours{} performs comparably to both the centralized and object-specialized baselines, which are retrained separately for each object. Across objects and metrics, no baseline consistently outperforms \ours{}. This suggests that the attachment-based interface provides object-level generalization without requiring either centralized joint-state control or per-object retraining.

\subsection{Cost of Transport}
\label{sec:cot}

We next examine how the energy efficiency of cooperative transport changes with team size. To account for differences in payload mass, we report Cost of Transport (CoT), a standard dimensionless measure of locomotion efficiency~\cite{collionscostoftransport}:
\begin{equation*}
    \text{CoT} = \frac{E}{m g d},
\end{equation*}
where $E$ is the total mechanical energy expended by the team during transport, $m$ is the payload mass, $g$ is gravitational acceleration, and $d$ is the horizontal distance traveled by the payload. Lower CoT indicates greater energy efficiency per unit payload weight and distance traveled.

\begin{figure}[t]
    \centering
    \includegraphics[width=0.9\columnwidth]{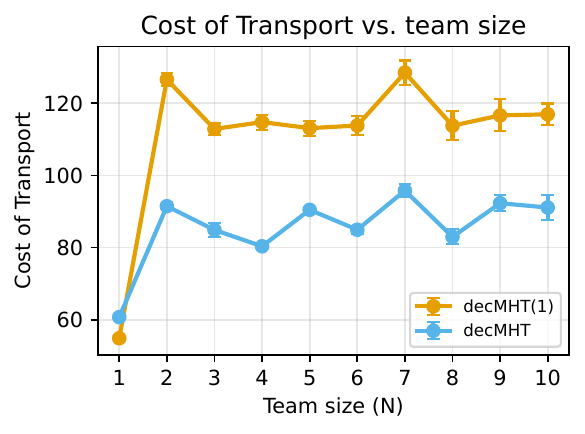}
    \vspace{-1em}
    \caption{Cost of Transport (CoT) as a function of team size with mass scaled linear with team size.}
    \label{fig:cot}
\end{figure}

Figure~\ref{fig:cot} reports CoT as a function of team size, comparing \ours{}(1) with \ours{}. As in Section~\ref{sec:scalability}, payload mass scales linearly with team size.
For both policy variants, CoT increases sharply from one to two robots and then largely plateaus as additional robots join the team. This suggests that cooperative transport incurs an initial coordination cost, but that this overhead does not continue to grow substantially with team size. Moreover, \ours{} maintains consistently lower CoT than \ours{}(1) for all multi-robot teams, despite being fine-tuned only with two robots. Thus, the efficiency gains from multi-robot training transfer zero-shot to substantially larger teams.

\subsection{Handover Evaluation}
\label{sec:handover_metrics}

We compare \ours{}(1) and \ours{} on three handover settings: \textbf{static}, where both robots remain stationary; \textbf{dynamic}, where the handoff occurs while the object is moving; and \textbf{cross-level}, where the robots stand at different elevations and the receiving robot must adjust its stance height.

Table~\ref{tab:handover_training} reports handover progress and peak object deviation for both policies. Robot $A$ initially holds the object and robot $B$ receives it. We measure success at three stages: \textbf{Stage 1}, $A$ holds the object alone; \textbf{Stage 2}, both robots hold it during the overlap; and \textbf{Stage 3}, $B$ holds it alone after the handoff. We also report peak height deviation and tilt during the handoff.

Across all three settings, \ours{} matches or improves on \ours{}(1), with the largest gains in the dynamic case, where $B$ must establish contact while the object is already moving. The strong cross-level performance is particularly notable because the policy is not trained with robots standing at different elevations.

\begin{table}[t]
    \caption{Handover performance across static, dynamic, and cross-level configurations.}
    \centering
    \footnotesize
    \setlength{\tabcolsep}{3pt}
    \label{tab:handover_training}
    \resizebox{\columnwidth}{!}{%
    \begin{tabular}{ll ccc cc}
        \toprule
        \shortstack{Handover\\config.} & Training & \shortstack{Stage 1\\(\%)} & \shortstack{Stage 2\\(\%)} & \shortstack{Stage 3\\(\%)} & \shortstack{Peak height\\dev.\ (m)} & \shortstack{Peak tilt\\(deg)} \\
        \midrule
        \multirow{2}{*}{Static}       & decMHT(1) & 100.0 & 100.0 & 97.0  & 0.063 & 8.98 \\
                                       & \ours{}  & 100.0 & 100.0 & 100.0 & 0.075 & 7.04 \\
        \midrule
        \multirow{2}{*}{Dynamic}      & decMHT(1) & 99.5  & 88.5 & 87.5 & 0.158 & 15.67 \\
                                       & \ours{}  & 100.0 & 98.0 & 96.5 & 0.119 & 8.49 \\
        \midrule
        \multirow{2}{*}{Cross-level} & decMHT(1) & 100.0 & 92.5 & 92.5 & 0.776 & 15.56 \\
                                       & \ours{}  & 100.0 & 99.0 & 99.0 & 0.776 & 14.61 \\
        \bottomrule
    \end{tabular}%
    }
\vspace{-1em}
\end{table}

\section{Sim-to-Real Transfer}
\label{sec:sim2real}

We validate \ours{} directly on hardware using two Digit V3 humanoids with no additional real-world fine-tuning. An external sensor provides the box's pose relative to each robot's own torso frame, giving both robots the attachment-frame pose ${}^{b}T^{(r)}$ required by their respective attachment specs. Using a cardboard box as the physical payload, we demonstrate all three operating regimes: (1) \textbf{single-robot pickup and transport}, with one robot approaching, lifting, and carrying the box under commanded planar velocity, yaw rate, and height; (2) \textbf{multi-robot pickup and transport}, with both robots simultaneously attached to opposite faces of the box and cooperatively tracking the same commanded object trajectory; and (3) \textbf{robot-to-robot handover}, in which one robot hands the box off to the other purely by toggling each robot's operating-mode flag, without any change to the policy, its inputs, or its training. See the supplementary video for example trials. 


\section{Limitations}

Our experiments focus on gripperless humanoids that maintain attachment through bimanual pinching forces. We expect the framework to extend naturally to robots with grippers, with the attachment spec providing local guidance for selecting and approaching grasp locations, but this has not yet been demonstrated. The current formulation also assumes that robots share a common behavioral role and differ primarily through their assigned attachment locations. Tasks requiring distinct roles, such as leader--follower behavior or specialized support and manipulation functions, would require extending the interface with robot-specific role information. In addition, each robot is assumed to track its attachment frame locally throughout execution. Our sim-to-real experiments obtain this information using an external motion capture. Replacing this external sensing with egocentric perception is an important direction for future work, particularly for deployment in unstructured environments.

\section{Summary}

We introduced \ours{}, a decentralized object-centric framework for multi-humanoid pickup, transport, and handover. Using only local attachment specs and transformed object-motion commands, a policy trained with at most two robots generalizes zero-shot to teams of up to ten humanoids, diverse payloads, and cross-level handovers, with sim-to-real transfer to two Digit V3 robots. To our knowledge, this is the first learned decentralized framework to demonstrate such capabilities across varying team sizes with robot-realistic humanoids and physical hardware.

\bibliographystyle{ieeetr}
\bibliography{references}

\end{document}